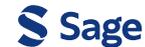

# Application of machine learning algorithms in classifying postoperative success in metabolic bariatric surgery: A comprehensive study

José Alberto Benítez-Andrades[1] ![ORCID], Camino Prada-García[2,3] ![ORCID], Rubén García-Fernández[4], María D Ballesteros-Pomar[5] ![ORCID], María-Inmaculada González-Alonso[4] and Antonio Serrano-García[6] ![ORCID]

## Abstract

**Objectives:** Metabolic bariatric surgery is a critical intervention for patients living with obesity and related health issues. Accurate classification and prediction of patient outcomes are vital for optimizing treatment strategies. This study presents a novel machine learning approach to classify patients in the context of metabolic bariatric surgery, providing insights into the efficacy of different models and variable types.

**Methods:** Various machine learning models, including Gaussian Naive Bayes, Complement Naive Bayes, K-nearest neighbour, Decision Tree, K-nearest neighbour with RandomOverSampler, and K-nearest neighbour with SMOTE, were applied to a dataset of 73 patients. The dataset, comprising psychometric, socioeconomic, and analytical variables, was analyzed to determine the most efficient predictive model. The study also explored the impact of different variable groupings and oversampling techniques.

**Results:** Experimental results indicate average accuracy values as high as 66.7% for the best model. Enhanced versions of K-nearest neighbour and Decision Tree, along with variations of K-nearest neighbour such as RandomOverSampler and SMOTE, yielded the best results.

**Conclusions:** The study unveils a promising avenue for classifying patients in the realm of metabolic bariatric surgery. The results underscore the importance of selecting appropriate variables and employing diverse approaches to achieve optimal performance. The developed system holds potential as a tool to assist healthcare professionals in decision-making, thereby enhancing metabolic bariatric surgery outcomes. These findings lay the groundwork for future collaboration between hospitals and healthcare entities to improve patient care through the utilization of machine learning algorithms. Moreover, the findings suggest room for improvement, potentially achievable with a larger dataset and careful parameter tuning.

[1]SALBIS Research Group, Department of Electric, Systems and Automatics Engineering, Universidad de León, León, Spain
[2]Department of Preventive Medicine and Public Health, University of Valladolid, Valladolid, Spain
[3]Dermatology Service, Complejo Asistencial Universitario de León, León, Spain
[4]Department of Electric, Systems and Automatics Engineering, Escuela de Ingenierías Industrial, Informática y Aeroespacial, Universidad de León, León, Spain

[5]Department of Endocrinology and Nutrition, Complejo Asistencial Universitario de León, León, Spain
[6]Psychiatry Service, Department of Psychosomatic, Complejo Asistencial Universitario de León, León, Spain

**Corresponding author:**
Camino Prada-García, Department of Preventive Medicine and Public Health, University of Valladolid, 47005 Valladolid, Spain.
Email:cprada@saludcastillayleon.es

## Introduction and related work

Metabolic bariatric surgery (MBS) has established itself as an effective intervention in the treatment of obesity, offering impressive weight loss results. However, outcomes are not universal and vary among patients and across surgical procedures.[1] Between 20% and 30% of patients experience suboptimal weight loss or significant weight regain (WR) within the first few postoperative years.[1]

Understanding the predictors of MBS outcomes is crucial for enhancing patient selection and postoperative interventions. Some studies have identified preoperative psychosocial factors such as depressive disorders and disordered eating patterns as negative predictors of weight loss outcomes.[2] Additionally, adherence to dietary and physical activity guidelines has emerged as positive predictors of weight loss.[2] Recent discussions and studies challenge the role of preoperative weight loss as a predictor of long-term postoperative outcomes, with some suggesting that it has been imposed more as a prerequisite by insurance companies in the US rather than being evidence-based.[3,4] Instead, emerging evidence indicates that early postoperative weight loss may serve as a more reliable predictor of maximum and long-term bariatric outcomes.[5-7] This shift in focus highlights the need for a nuanced understanding of weight loss dynamics in MBS.

Despite the effectiveness of MBS, there is substantial variability in long-term weight loss outcomes, and few factors have predictive power to explain this variability. Neuroimaging may provide a novel biomarker with utility beyond other commonly used variables in MBS trials to improve the prediction of long-term weight loss outcomes.[8] Predicting responses to bariatric and metabolic surgery has become an essential tool in clinical practice, and various pre-operative factors have been proposed as predictors of body weight reduction after surgery.[9]

Some patients experience WR or insufficient weight loss after MBS, and the mechanisms and preoperative predictors contributing to WR are complex and multifactorial.[10] Identifying factors related to weight loss maintenance in the medium and long term after MBS is vital to reduce failure.[11] Quality of life outcomes, including weight-related quality of life, have also been identified as important patient-reported outcomes for obesity treatment.[12] Therefore, MBS is a valuable tool in the management of obesity, but variability in outcomes underscores the need for further research into pre and postoperative predictors. Identifying these factors may lead to more targeted interventions and improved quality of life for patients undergoing MBS.

Additionally, recent studies like the one conducted by Saux et al.[13] have underscored the significance of conventional predictors in MBS outcomes, including physiological measures such as type of surgery, preoperative weight, and height. While acknowledging the critical insights provided by these conventional predictors, our study aims to broaden the predictive landscape by incorporating a robust set of psychosocial variables. These variables, which focus on psychological well-being, social support, and behavioural patterns, seek to add depth and nuance to the understanding of patient outcomes post-surgery. They represent factors often underrepresented in surgical outcome predictions but suggested by evidence to significantly impact short-term health outcomes. In doing so, we provide a more comprehensive model that encompasses both the established physiological factors and the emerging psychosocial factors, aiming for a holistic understanding of the patient's journey through MBS.

However, various artificial intelligence (AI) techniques, including machine learning among others, are increasingly being used in the field of medical diagnosis and treatment. For example, AI-based algorithms have been employed to predict acute kidney injury after cardiac surgery, utilizing techniques like logistic regression, support vector machine, random forest (RF), and a combined model (RF + XGboost).[14] By using AI-based systems, researchers can collaborate in real-time and share knowledge digitally to potentially heal millions.

In the context of spine surgery, AI-driven prediction modelling using hybrid machine learning models has been explored. The techniques discussed could become important in establishing a new approach to decision-making in spine surgery based on three fundamental pillars: Patient-specific, AI-driven, and integrating multimodal data.[15]

AI and machine learning have also been applied in orthopaedic surgery, where a systematic review protocol aims to provide an update on AI advancements in clinical diagnosis and prediction of postoperative outcomes and complications.[16]

In the challenging environment of the coronavirus disease 2019 (COVID-19) pandemic, AI and machine learning approaches have been employed for mortality risk prediction in surgical patients with perioperative severe acute respiratory syndrome related coronavirus 2. Patient factors, rather than operation factors, were found to be the best predictors.[17]

While these studies demonstrate the growing application of AI in various surgical fields, the specific application of AI in MBS remains an area of potential exploration and innovation.[18-20] The success of AI in predicting diagnoses in other medical domains suggests that similar methodologies could be applied to MBS to enhance predictive accuracy, optimize treatment strategies, and improve patient outcomes.

This research explores the application of various machine learning models, including Gaussian Naive Bayes (GaussianNB), Complement Naive Bayes (ComplementNB), K-nearest neighbour (KNN), Decision Tree, KNN with RandomOverSampler, and KNN with SMOTE, on a dataset of 73 patients who have undergone MBS. The dataset, provided by the Complejo Asistencial Universitario de León and converted from Excel to CSV format, consists of 70 variables, encompassing psychometric, socioeconomic, and analytical characteristics of the patients. The models were applied to these variable groupings to determine the most efficient predictive approach.

The paper is structured as follows: The 'Methodology' section delineates the machine learning models employed, details the hyperparameter adjustments, and elucidates the oversampling techniques utilized. It also provides the rationale behind the selection of specific variable groupings. The 'Experiments and Results' section offers a comprehensive description of the dataset and presents the experimental setup and findings. Finally, the 'Discussion' section delves into the interpretation of the findings, their broader implications, and potential limitations. The 'Conclusions' section summarizes the key takeaways of the research and outlines future research avenues and potential improvements.

## Methodology

The methodology used to carry out this research, from data collection to modelling results, is summarized in Figure 1.

### Traditional techniques

To identify the most impactful variables in MBS, we employed several machine learning algorithms, including Logistic Regression, GaussianNB, ComplementNB, KNN, and decision trees. Additionally, we conducted experiments using oversampling methods like RandomOver and Smote. In this part, we will present the foundational versions of the techniques applied. Here, we provide a rationale for the selection of each model, supported by scientific references pertinent to our case of MBS:

- *Logistic Regression:* A fundamental tool for binary classification problems, Logistic Regression offers a baseline for interpretability and performance comparison. It is widely used in medical research due to its ability to provide odds ratios, making the influence of predictors on outcomes explicit.[21]
- *Naive Bayes:* Naive Bayes is included for its efficiency and ease of implementation, particularly valuable in early stages of data analysis. Its effectiveness is well-documented across various dataset sizes, demonstrating adaptability and performance. Importantly, it has shown success in studies with similar sample sizes to ours, exemplifying its applicability in settings akin to our research. This is particularly evidenced in studies like the one by Dominguez-Rodriguez et al., where Naive Bayes was effectively employed in scenarios with comparable sample sizes, underlining its utility in complex classification tasks. The model's robust performance in such diverse and often imbalanced clinical datasets validates its inclusion, ensuring a well-supported and reliable approach for our study.[22,23]
- *K-nearest Neighbours*: KNN is included due to its capability to handle complex, non-linear interactions

between variables without a priori assumptions about data distribution. It has shown effectiveness in various medical classification tasks, For example, KNN has been found to outperform other classification techniques in diagnosing heart diseases, emphasizing its suitability for complex medical classification scenarios.[24]
- *Decision Trees:* Decision Trees offer a visual and interpretable way to represent decision-making processes. Their ability to capture non-linear relationships and interactions between variables makes them suitable for medical decision-making contexts. They have been reported to have a high accuracy rate in identifying students' problems during COVID-19, for example, with a 95.85% accuracy rate.[25]
- *Oversampling Methods (RandomOver and SMOTE):* In addressing the prevalent class imbalance in medical datasets, we employed RandomOver for its straightforward replication approach to augment minority classes, and SMOTE for generating synthetic yet plausible samples through interpolation. These methods have been crucial in improving classifier performance by enriching the diversity and representativeness of training data. Empirical studies confirm that both techniques significantly enhance predictive accuracy and robustness in varied medical contexts, effectively addressing class imbalance challenges[26]). Their combined usage provides a comprehensive strategy to improve model reliability and outcome prediction in imbalanced datasets.

Each model was selected to provide a broad perspective on variable importance in MBS outcomes, ensuring a comprehensive analysis through diverse methodological lenses. The references cited offer empirical and theoretical support for the use of these methods in similar medical contexts, reinforcing the applicability and relevance of our approach.

*Logistic regression.* Logistic Regression is a predictive analysis used primarily for binary classification. It models the probability of a default class (usually the '1' class in binary classification) using the logistic function. The logistic function ensures output values between 0 and 1, representing the probability of occurrence of one of the classes.

The probability that an input vector $X$ belongs to the default class '1' is given by:

$$P(Y = 1|X) = \frac{1}{1 + e^{-(\beta_0 + \beta_1 X_1 + \cdots + \beta_k X_k)}} \quad (1)$$

Here, $X_1, X_2, \ldots, X_k$ are the explanatory variables, $\beta_0, \beta_1, \ldots, \beta_k$ are the coefficients, and $e$ is the base of the natural logarithm. The model is fitted using techniques like maximum likelihood estimation.

Despite its simplicity, Logistic regression is a powerful method for binary classification problems, especially when the log-odds of the dependent variable are a linear combination of the independent variables. It's widely used due to its efficiency and interpretability.

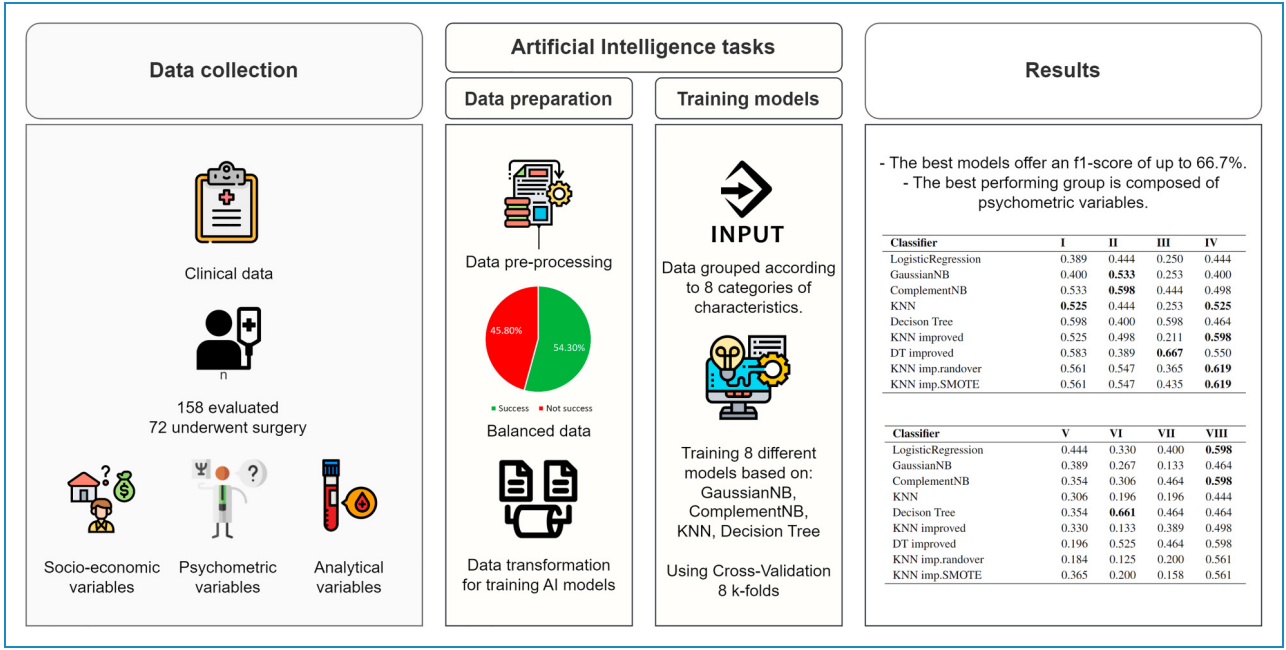

**Figure 1.** Methodology applied in research from the collection of data from metabolic bariatric surgery patients to obtaining predictive models of success in metabolic bariatric surgery.

*Gaussian Naive Bayes.* The GaussianNB classifier builds upon the Naive Bayes algorithm, which itself is grounded in the Bayes Theorem for computing conditional probabilities.

Given $P(A)$ as the likelihood of event $A$; $P(B)$ as the likelihood of event $B$; $P(A|B)$ as the likelihood of event $A$ given $B$; $P(B|A)$ as the likelihood of event $B$ given $A$, and $P(A \cap B)$ as the likelihood of both events $A$ and $B$ happening, the Bayes Theorem for conditional probability calculation is expressed by equation 2.

$$P(A|B) = \frac{P(B|A) \times P(A)}{P(B)} \qquad (2)$$

The classic Naive Bayes classifier requires the variables to be completely independent. Despite this, Naive Bayes is favoured for its simplicity and minimal training data requirements for effective results.

In our study, we utilized GaussianNB, a variant that accommodates continuous data by assuming a normal (Gaussian) distribution, preserving the benefits of the original method. The GaussianNB conditional probability is determined by the equation below:

$$P(x) = \frac{1}{\sqrt{2\pi\sigma^2}} e^{-\frac{(x-\mu)^2}{2\sigma^2}} \qquad (3)$$

where $\mu$ is the mean, computed by the equation below:

$$\mu = \frac{1}{n}\sum_{i=1}^{n} x_i \qquad (4)$$

and $\sigma$ is the Standard deviation, computed by the equation below:

$$\sigma = \sqrt{\frac{1}{n-1}\sum_{i=1}^{n}(x_i - \mu)^2} \qquad (5)$$

*Complement Naive Bayes.* This derivative of the Naive Bayes algorithm is typically employed with datasets that have imbalanced data.

It addresses the imbalance by calculating the likelihood of an element belonging to all classes rather than a specific class. The process involves:

1. Calculating the probability that the data does not belong to each class.
2. Selecting the smallest value from those obtained.
3. Choosing this lowest probability, indicating the class to which the element does not belong. This unique approach gives the method its name, ComplementNB.

*K-nearest neighbours.* KNN is a renowned method that classifies input data based on the distances between that data and other elements within each class. Specifically, it selects the K nearest neighbours to the input data, classifying the input within the class that the majority of the selected K neighbours belong to.

The distance is commonly calculated using the Euclidean distance, as shown in the equation below:

$$d(x, x') = \sqrt{(x_1 - x'_1)^2 + \cdots + (x_n - x'_n)^2} \quad (6)$$

After determining the distances, the input $x$ is assigned to the class with the highest likelihood (7).

$$P(y = j | X = x) = \frac{1}{K} \sum_{i \in A} I(y^i = j) \quad (7)$$

Selecting the optimal value for parameter k requires consideration of the problem type, data type, and other factors, making grid search essential for choosing the best value.

*Decision tree.* Decision trees consist of the root node, representing all data, decision nodes, and leaf nodes where no further splits occur. The algorithm considers two key concepts for data splitting: Entropy and the Gini index.

Entropy measures the information needed to describe an element, as given by the equation below.

$$E = -\sum_{i=1}^{n} p_i * log(p_i) \quad (8)$$

The Gini index reflects data homogeneity, with 0 indicating total homogeneity and 1 indicating total heterogeneity, as shown in the equation below.

$$G = 1 - \sum_{i=1}^{n} p_i^2 \quad (9)$$

The algorithm's steps include selecting the root node based on low entropy and high information gain, then iteratively calculating entropy and information gain for unused nodes, selecting the base node with the least entropy and greatest information gain, and repeating this process until the tree is complete.

*Oversampling.* Oversampling methods like RandomOver and SMOTE are employed to enhance methods and results on unbalanced datasets.[27,28] Balancing can be achieved by either deleting data from the majority class or duplicating data from the minority class. In this study, we utilized RandomOver and SMOTE as oversampling techniques:

- RandomOver: This straightforward method randomly selects and duplicates samples from the minority class until balance is achieved.
- SMOTE: Unlike duplication, SMOTE[29] creates synthetic data similar to the minority class. It randomly selects a sample, calculates the k closest neighbours (usually 5), and generates new data using a distance metric to compute differences between the feature vector and selected neighbours, multiplied by a random value between 0 and 1.

## Experiments and results

### Dataset

The dataset was obtained after a retrospective revision of the clinical data of all patients who underwent Metabolic and Bariatric Surgery (MBS) and followed up for at least one year and completed the pre-surgical assessment process. A total of 158 patients were evaluated by a multidisciplinary team, including specialists in endocrinology, gastroenterology, general surgery, and psychiatry. Out of these, 72 patients met the inclusion criteria, underwent surgery, and completed a one-year follow-up. Clinical and analytical variables were collected in the Endocrinology and Nutrition and Psychiatry departments following the MBS protocol approved by the Centre.

The types of MBS performed on patients included one or a combination of the following: Sleeve Gastrectomy (Gastrectomía Tubular), Biliopancreatic Diversion, or Single Anastomosis Duodeno-Ileal with Sleeve. These procedures were selected based on individual patient assessments and surgical indications.

The criteria to be included in the MBS process were:

1. Excess weight with a body mass index (BMI) greater than 40 or a BMI greater than 35 and associated medical problems.[30]
2. Age over 18 years.
3. Absence of contraindication due to digestive or psychiatric pathologies.
4. Indication by the surgeon of the surgical technique.

The criteria to define the success of the process corresponds to a loss of at least 50% of the excess weight, representing the ideal weight as a BMI of 25.

The variables gathered from each patient are categorized into three distinct groups. These groups, along with the validated questionnaires used for collection, are listed below:

1. *Socio-economic variables:* This group is made up of the variables gender, age, employment status and educational level.
2. *Psychometric variables:* In this group we found the variables of existence of personal or family psychiatric history and the score of different scales and subscales, as well as the existence of any current psychiatric disorder that does not contraindicate the performance of the procedure:
   EuroQoL5 Quality of Life Scale[31]: (i) Mobility, (ii) Need for care, (iii) Activity, (iv) Pain, (v) Depression, (vi) Perceived quality of life scale (VAS).
   Salamanca Screening Questionnaire[32]: (i) Paranoid trait, (ii) Schizoid trait, (iii) Schizotypal trait, (iv) Histrionic trait, (v) Antisocial trait, (vi) Narcissistic trait, (vii) Impulsive trait, (viii) Borderline trait, (ix) Anankastic trait, (x) Dependent trait, (xi) Anxious trait. Edinburgh Bulimia Scale[33]: (i) Symptom subscale, (ii) Severity subscale,

ACTA Inventory of motivation for change[34]: (i) Precontemplative stage, (iv) Contemplative stage, (v) Decision stage, (vi) Action stage, (vii) Maintenance stage, (viii) Relapse stage.

Plutchik's Wheel of Emotions.[35]

3. *Analytical variables:* These are variables from the patients' blood tests: (i) Haemoglobin (Hb), (ii) Haematocrit (Hct), (iii) Basal glucose, (iv) Glycosylated haemoglobin (HbA1c), (v) Basal insulin, (vi) Cholesterol, (vii) Triglycerides, (viii) High-Density Lipoprotein (HDL) cholesterol, (ix) Uric acid, (x) Glutamic-Oxaloacetic Transaminase (GOT), also known as Aspartate Aminotransferase (AST), (xi) Glutamic Pyruvic Transaminase (GPT), also known as Alanine Aminotransferase (ALT), (xii) Gamma-Glutamyl Transferase (GGT), (xiii) Iron, (xiv) Ferritin, (xv) Transferrin, (xvi) Albumin, (xvii) Prealbumin, (xviii) Phosphorus, (xix) Parathyroid hormone (PTH), (xx) Osteocalcin, (xxi) Crosslaps, (xxii) Vitamin D, (xxiii) C-reactive protein (CRP).

## Experimental setup

In the initial phase, the data underwent preprocessing to ensure they were clean and formatted appropriately for analysis. The input variables fall under three categories previously mentioned: socio-economic, psychometric, and analytical. The outcome variable is termed 'success', indicating whether the MBS was successful (value 1) or not (value 0). Out of the 73 patients, 26.4% were female and 73.6% were male. In terms of the success rate, 45.8% were unsuccessful, while 54.2% were successful.

**Table 1.** Division of input variables into groups.

| Group | Description |
| --- | --- |
| Group I | Socio-economic variables |
| Group II | Psychometric variables |
| Group III | Analytical variables |
| Group IV | Only psychometric variables from the EuroQoL5 Quality of Life Scale |
| Group V | Only the psychometric variables from the Salamanca screening questionnaire |
| Group VI | Only the psychometric variables from the ACTA Inventory of motivation for change |
| Group VII | Combination of all variables (Groups I, II, and III) |
| Group VIII | Combination of socio-economic variables (Group I) and psychometric variables (Group II) |

Pipelines were generated to convert the categorical variables using a LabelEncoder and the continuous variables were scaled. After this, a selection of best features was performed using SelectKBest and ExtraTrees.

For the experiments, the input variables were segmented into 8 distinct groups showed in Table 1:

The experiments consisted of applying the following models to predict the outcome variable. The nomenclature of the models used is detailed in Nomenclature section.

A k-fold cross-validation with 8 folds was conducted for all experiments to mitigate randomization. Each model was assessed using the most suitable hyperparameter selection techniques based on the input data they were applied to. Furthermore, RandomOver and SMOTE oversampling techniques were employed in some instances.

## Results

Figure 2 shows a graph representing the results obtained by the 8 aforementioned classification models, applied to each of the outcome variable to be predicted using the eight aforementioned groups of variables. In addition, Table 2 shows the results of all of the experiments carried out in numerical values.

Two particularly high values are observed, corresponding to a 66.1% success rate in the decision trees on the group VI variable (ACTA psychometrics) and 66.7% success rate on improved decision trees (with hyperparameter tuning) applied to group III (analytical variables).

By grouping the f1-score metrics obtained for each group of variables in all the experiments carried out, the means and standard deviations shown in the Table 3 were obtained. The group of variables with the best mean f1-score metric was the group formed by the socio-economic variables together with the psychometric variables. This group consisted of the grouping of both groups described in the Dataset subsection.

## Discussion

Following the experiments detailed in the previous section, socioeconomic variables had a significant impact on the predictive models, with an average f1-score of 0.532. Psychometric variables, which include factors such as personal or family psychiatric history and scores from different scales and subscales, achieved an average f1-score of 0.489. On the other hand, analytical variables, referring to blood test results, had an average f1-score of 0.403.

When specific psychometric variables were analyzed, the EuroQol5 quality of life scale performed best with an average f1-score of 0.524. In contrast, the Salamanca screening questionnaire and the ACTA motivation for change inventory had average f1-scores of 0.325 and 0.305, respectively. Combining all variables (socioeconomic, psychometric, and analytical) resulted in an average f1-score of 0.319. However, the combination of socioeconomic and psychometric variables yielded the highest average f1-score of 0.532.

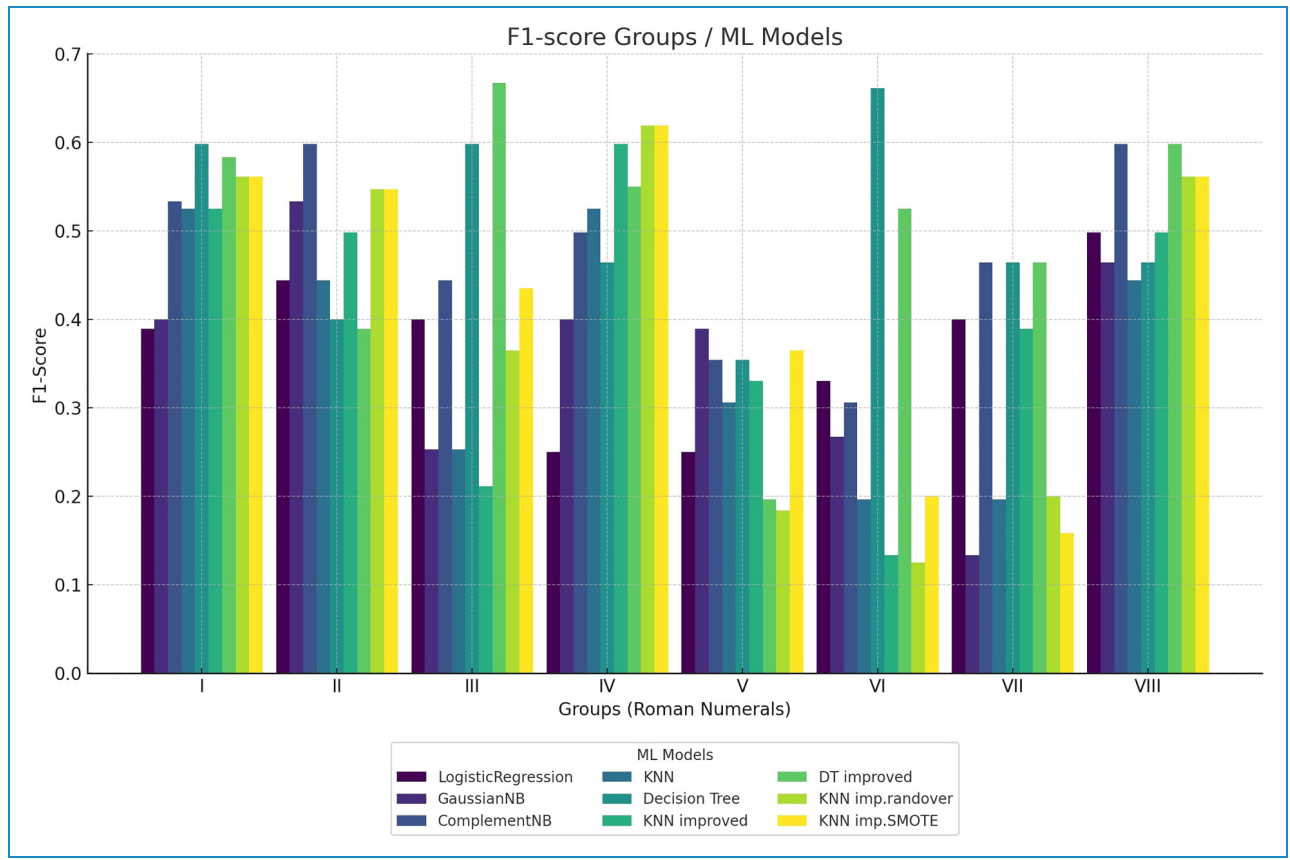

**Figure 2.** Graph representing the trained classification models and the f1-score obtained in each of them.

The findings from our study underscore the significance of socioeconomic and psychometric variables in predicting the success of MBS. These insights align with previous research that has emphasized the relevance of non-clinical factors in predicting outcomes in MBS. For instance, a study by Chevallier et al.[36] highlights that factors such as age, initial BMI, surgeon's experience, recovery of physical activity, and change in eating habits are predictive of the success of MBS. Additionally, studies like those by Gordon et al.[37] and Belanger et al.[38] have demonstrated that psychosocial and psychometric variables assessed during preoperative evaluation significantly predict weight loss after MBS, with a more substantial impact observed in long-term follow-up. These findings highlight the multifactorial complexity in predicting the success of MBS, where integrating psychological and socioeconomic variables along with clinical indicators provides a more comprehensive perspective to enhance postoperative outcomes and patient quality of life.

The observation that socioeconomic variables in our research had a notable impact on the predictive models,

with an average f1-score of 0.519, is particularly intriguing. Cummins et al.[39] also discovered that socioeconomic and demographic characteristics can influence the type of MBS performed in adolescents. While their research focused on procedure choice rather than success prediction, it emphasizes the importance of considering socioeconomic factors in bariatric research and practice.

In terms of psychometric variables, our study identified that the EuroQol5 quality of life scale obtained the best results, with a mean f1 score of 0.524. Furthermore, individually, of the 7 models trained, this group of variables scored the best in 4 of them, ranging from 0.525 to 0.619. This suggests that patient-perceived quality of life may be a key indicator of the success of MBS. Although no directly comparable study was found in the literature reviewed, the importance of psychometric variables in predicting medical outcomes has been recognized in other contexts, such as in the study by Bhanderi et al.[40] which investigated the influence of socioeconomic deprivation on the performance of MBS.

Referencing the results of each model by group, without considering the average, the group that achieved the best result was the analytical variables with the DT Improved model, scoring an F1-score of 0.667. This was followed by the psychometric variables from ACTA with an F1-score of 0.661 in the DT model. In third place were the psychometric variables

**Table 2.** Values of f1-score obtained in each classification model for each of the 8 groups of input variables used.

| Classifier | I | II | III | IV | V | VI | VII | VIII |
|---|---|---|---|---|---|---|---|---|
| LogisticRegression | 0.389 | 0.444 | 0.250 | 0.444 | 0.444 | 0.330 | 0.400 | **0.598** |
| GaussianNB | 0.400 | **0.533** | 0.253 | 0.400 | 0.389 | 0.267 | 0.133 | 0.464 |
| ComplementNB | 0.533 | **0.598** | 0.444 | 0.498 | 0.354 | 0.306 | 0.464 | **0.598** |
| KNN | **0.525** | 0.444 | 0.253 | **0.525** | 0.306 | 0.196 | 0.196 | 0.444 |
| DT | 0.598 | 0.400 | 0.598 | 0.464 | 0.354 | **0.661** | 0.464 | 0.464 |
| KNN improved | 0.525 | 0.498 | 0.211 | **0.598** | 0.330 | 0.133 | 0.389 | 0.498 |
| DT improved | 0.583 | 0.389 | **0.667** | 0.550 | 0.196 | 0.525 | 0.464 | 0.598 |
| KNN imp.randover | 0.561 | 0.547 | 0.365 | **0.619** | 0.184 | 0.125 | 0.200 | 0.561 |
| KNN imp.SMOTE | 0.561 | 0.547 | 0.435 | **0.619** | 0.365 | 0.200 | 0.158 | 0.561 |

GaussianNB: Gaussian Naive Bayes; ComplementNB: Complement Naive Bayes; KNN: k-Nearest Neighbor; DT: Decison Tree.
Bold text represents the highest accuracy obtained by each classifier.

**Table 3.** Mean F1-score and standard deviation (SD) obtained according to the group of variables used in all experiments.

| Variable Group | Mean | SD |
|---|---|---|
| Socio-economic | 0.519 | 0.071 |
| Psychometric | 0.489 | 0.069 |
| Analytical | 0.386 | 0.154 |
| EuroQol.5 | 0.524 | 0.075 |
| Salamanca screening | 0.325 | 0.081 |
| ACTA | 0.305 | 0.171 |
| All variables | 0.319 | 0.135 |
| Socio-economic + Psychometric | **0.532** | 0.060 |

Bold text represents the highest mean accuracy value of all variable groups used as input.

from the EuroQol.5, which scored 0.619 in the KNN models with oversampling (SMOTE and RandoverSampler). It is worth noting that although the analytical variables achieved the highest F1-score in one of the 7 trained models, their average F1-score is significantly lower at 0.403, compared to the F1-score achieved by models trained with the EuroQol.5 variable group. A similar trend is observed with the ACTA variables, which had an average F1-score of 0.301.

Lastly, the amalgamation of socioeconomic and psychometric variables in our research resulted in the highest average f1-score of 0.532. This indicates that a combined approach might be the most effective in predicting the success of MBS. Although the reviewed literature did not provide a directly comparable study on this aspect, the notion of integrating multiple variables to enhance predictive.

## Conclusions

Our study contributes valuable insights into the importance of socioeconomic and psychometric variables in predicting the success of MBS. These findings underscore the need to consider a broad range of factors when assessing and planning bariatric interventions, particularly those related to the patient's socioeconomic status and psychological profile.

The main limitations of our study arise from its retrospective design and the inherent constraints of the dataset used. From an initial pool of 158 patients, only 72 met the inclusion criteria and completed a one-year follow-up. This sample might not be representative of the broader population undergoing MBS. The dataset, sourced from a non-randomized clinical review, could introduce biases not found in prospective or randomized studies. It's crucial to emphasize the need for more comprehensive prospective research to validate our findings and guide interventions in the pre-surgical assessment of MBS candidates. An additional bias to note is that patients with psychopathology were excluded from the procedure, leaving out more altered psychometric results from the analysis.

Despite these limitations, the importance of our study is clear. While acknowledging the small sample size, it is essential to recognize that similar research has provided valuable insights into specialized medical fields. Studies such as[41,42,22] and as indicated by Vabalas et al.,[43] demonstrate that, with rigorous validation techniques and methodological designs, even small datasets can lead to robust and impactful conclusions. Our study, focusing on a specific patient population and integrating detailed psychometric and socioeconomic variables, aligns with this precedent, emphasizing the potential to uncover meaningful insights even from smaller samples. Such work is increasingly vital in medical fields where large-scale data may not be available, and where ethical and practical considerations necessitate innovative approaches to research.

In conclusion, while recognizing the need for further research to validate and extend our findings, the current study adds to the growing body of evidence that smaller, well-designed studies are not only feasible but also crucial in advancing our understanding of complex medical phenomena. Future studies should aim to expand on this work, incorporating larger and more diverse patient samples and employing even more rigorous methodological approaches.

**Contributorship:** José Alberto Benítez-Andrades contributed to the conceptualization, data curation, methodology, software, visualization, validation, supervision, writing - original draft preparation. **Camino Prada-García** contributed to the conceptualization, visualization, validation, writing – original draft preparation. **Rubén Garcí-Fernández** contributed to the conceptualization, data curation, methodology, software, visualization, writing – reviewing and editing. **María D. Ballesteros-Pomar** contributed to the conceptualization, supervision, writing – reviewing and editing. **Inmaculada González-Alonso** contributed to the conceptualization, supervision, writing – reviewing and editing. **Antonio Serrano-Garía** contributed to the conceptualization, data curation, methodology, validation, supervision, writing – original draft preparation.

**Declaration of conflicting interests:** The authors declared no potential conflicts of interest with respect to the research, authorship, and/or publication of this article.

**Ethical approval:** According to the protocol approved by the Ethics Committee for Research with Medicines of the León and Bierzo Health Areas, informed consent is not required for this study. This study was approved by the Ethics Committee of the León and Bierzo Health Areas on 29 November 2022.

**Funding:** The authors received no financial support for the research, authorship, and/or publication of this article

**Guarantor:** José Alberto Benítez-Andrades.

**ORCID iDs:** José Alberto Benítez-Andrades  https://orcid.org/0000-0002-4450-349X
María D Ballesteros-Pomar  https://orcid.org/0000-0002-5729-9926
Camino Prada-García  https://orcid.org/0000-0002-2487-9350
Antonio Serrano-García  https://orcid.org/0000-0003-0704-3671

## Nomenclature and description of models trained for the experiments.

| | |
|---|---|
| LR | Logistic regression. |
| GaussianNB | Gaussian Naive Bayes. |
| ComplementNB | Complement Naive Bayes. |
| KNN | k-nearest neighbours. |
| DT | Decision tree. |
| KNN improved | KNN with optimized hyperparameters. |
| DT improved | Decision tree with optimized hyperparameters. |
| KNN imp.randover | KNN with optimized hyperparameters and RandomOver. |
| KNN imp.SMOTE | KNN with optimized hyperparameters and SMOTE. |